\documentclass[letterpaper]{article} 
\usepackage{aaai23}  
\usepackage{times}  
\usepackage{helvet}  
\usepackage{courier}  
\usepackage[hyphens]{url}  
\usepackage{graphicx} 
\usepackage{natbib}  
\usepackage{caption} 
\usepackage{algorithm}
\usepackage{algorithmic}

\usepackage{subcaption}
\usepackage{caption}
\usepackage{epsfig}
\usepackage{bm}
\usepackage{makecell}
\usepackage{multirow}
\usepackage{adjustbox}
\usepackage{bbm}
\usepackage{epstopdf}
\usepackage{threeparttable}
\usepackage{tablefootnote}

\usepackage{amsmath,amssymb}
\usepackage{graphicx}
\usepackage{color}
\usepackage{bm}
\usepackage{algorithm}
\usepackage{algorithmic}
\usepackage{url}

\newcommand{\benr}{\begin{eqnarray}}
\newcommand{\eenr}{\end{eqnarray}}
\newcommand{\benrr}{\begin{eqnarray*}}
\newcommand{\eenrr}{\end{eqnarray*}}
\newcommand{\ben}{\begin{equation}}
\newcommand{\een}{\end{equation}}
\newcommand{\benn}{\begin{equation*}}
\newcommand{\eenn}{\end{equation*}}

\newcommand{\cL}{\mathcal L}

\newcommand{\cX}{\mathcal X}
\newcommand{\cY}{\mathcal Y}

\newcommand{\R}{\mathbb R}

\usepackage{newfloat}
\usepackage{listings}
\DeclareCaptionStyle{ruled}{labelfont=normalfont,labelsep=colon,strut=off} 
\floatstyle{ruled}
\newfloat{listing}{tb}{lst}{}
\floatname{listing}{Listing}
\DeclareMathAlphabet{\mathcal}{OMS}{cmsy}{m}{n}

\title{Fair-CDA: Continuous and Directional Augmentation for Group Fairness}

\author{
    Rui Sun\textsuperscript{\rm 1,\rm 2}\equalcontrib, Fengwei Zhou\textsuperscript{\rm 3}\equalcontrib, Zhenhua Dong\textsuperscript{\rm 3},
    Chuanlong Xie\textsuperscript{\rm 4}\footnote{Correspondence to: lizhen@cuhk.edu.cn,clxie@bnu.edu.cn},
    Lanqing Hong\textsuperscript{\rm 3},
    Jiawei Li\textsuperscript{\rm 3}, \\
    Rui Zhang\textsuperscript{\rm 5},
    Zhen Li\textsuperscript{\rm 1,2}\footnotemark[2],
    Zhenguo Li\textsuperscript{\rm 3}
}

\affiliations {
    \textsuperscript{\rm 1} The Future Network of Intelligence Institute, The Chinese University of Hong Kong (Shenzhen) \\
    \textsuperscript{\rm 2} School of Science and Engineering, The Chinese University of Hong Kong (Shenzhen)\\
    \textsuperscript{\rm 3} Huawei Noah's Ark Lab\\
    \textsuperscript{\rm 4} Beijing Normal University\\
    \textsuperscript{\rm 5} Tsinghua University \\
   \ ruisun@link.cuhk.edu.cn, fzhou@connect.ust.hk, \{dongzhenhua, honglanqing, li.jiawei, li.zhenguo\}@huawei.com, rayteam@yeah.net, clxie@bnu.edu.cn, lizhen@cuhk.edu.cn \\
}

\usepackage{bibentry}

\begin{document}

\maketitle

\begin{abstract}
In this work, we propose {\it Fair-CDA}, a fine-grained data augmentation strategy for imposing fairness constraints. We use a feature disentanglement method to extract the features highly related to the sensitive attributes. Then we show that group fairness can be achieved by regularizing the models on transition paths of sensitive features between groups. By adjusting the perturbation strength  in the direction of the paths, our proposed augmentation is controllable and auditable. To alleviate the accuracy degradation caused by fairness constraints,  we further introduce a calibrated model to impute labels for the augmented data. Our proposed method does not assume any data generative model and ensures good generalization for both accuracy and fairness. Experimental results show that Fair-CDA consistently outperforms state-of-the-art methods on widely-used benchmarks, e.g., Adult, CelebA and MovieLens. 
Especially, Fair-CDA obtains an 86.3\% relative improvement for fairness while maintaining the accuracy on the Adult dataset.
Moreover, we evaluate Fair-CDA in an online recommendation system to demonstrate the effectiveness of our method in terms of accuracy and fairness.
\end{abstract}

\section{Introduction}

Many machine learning systems have achieved empirically success in practical problems but may sometimes raise issues of discrimination and unfairness.
In job candidate search, different protected groups (e.g., gender and ethnic groups) may be treated unfairly in terms of their members appearing in recommended candidate lists~\cite{ekstrand2021fairness}.
In the context of information retrieval, unfairness may happen among multiple parties. For example, unfair exposure allocation may favour monopolies and drive small content providers out of the market~\cite{Morik_2020}. This reduces diversity and impairs the whole ecosystem.

There have been various studies to impose fairness constraints during training procedure~\cite{Zemellearningfair,hardt2016equality,zafar2017fairness,chuang2021fair}, ensuring that different groups shall be treated similarly.
However, these constraints are data-dependent, the learnt fair classifiers might not generalize at evaluation time.
\citet{agarwal2018reductions} and \citet{cotter2019training} consider two-player games to formulate the constrained optimization problem and analyze the solutions and generalization guarantees.
\citet{chuang2021fair} proposes {\it Fair Mixup} to
generate a path of distributions that connects sensitive groups and regularize the smoothness of transitions among the path to improve the generalization of group fairness metrics. 
They show that their strategy ensures a better generalization for both accuracy and fairness in a wide range of benchmarks. 

\begin{figure*}[t]
\centering
\includegraphics[width=0.9\textwidth]{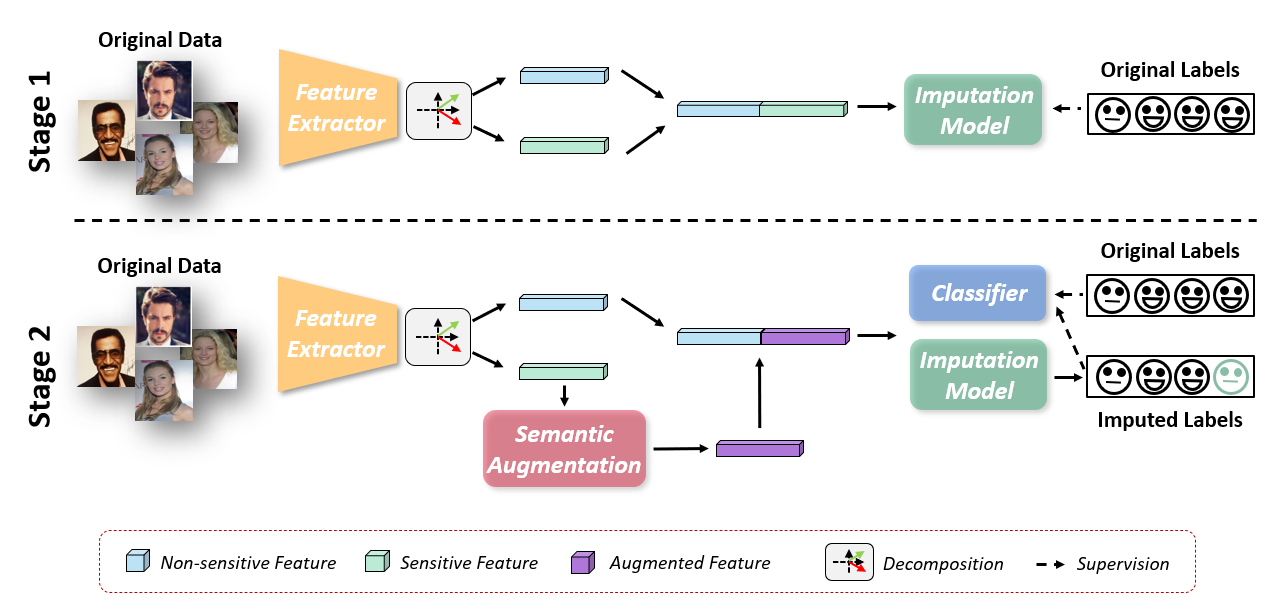}
\caption{An overview of the proposed Fair-CDA.}
\label{fig:model}
\end{figure*}

Motivated by {\it Fair Mixup}, we propose {\it Fair-CDA}, a continuous and directional augmentation method, to seek a fine-grained balance between fairness and accuracy. An overview of our method is illustrated in Figure~\ref{fig:model}.

\noindent{\bf Accuracy.} The {\it Mixup} \cite{zhang2018mixup} generates augmented samples via convex combinations of pairs
of data points. However, the between-group augmentation performs {\it Mixup} on both sensitive attributes and non-sensitive attributes.
This may change the correlation between non-sensitive attributes and the targets and further lead to the fall of prediction accuracy.
What's more, \citet{verma2019manifold} shows that interpolations in deeper hidden layers, which capture higher-level information \cite{zeiler2014visualizing}, can provide additional training signal and smooth decision boundaries that benefit generalization.
Therefore we develop a fine-grained augmentation via feature disentanglement and focus on the transitions over sensitive features.

We decompose representations in latent space into sensitive and non-sensitive features via {\it DecAug} \cite{bai2020decaug}, which is a powerful feature disentanglement technique for  Out-of-Distribution (OoD) generalization. 
The sensitive features encode the information that is strongly correlated to the sensitive attributes, while the non-sensitive features retain as much other information (essential for prediction) as possible.
Then we apply semantic augmentation on the sensitive features aiming to generate features correlated to the opposite sensitive attributes. The augmented sensitive features combined with the original non-sensitive features form what we refer to as the augmented features.

\noindent{\bf Fairness.}  We eliminate the disparities of the predictions made by a task model via training the model to make the same decisions for the samples with the original features and with the corresponding augmented features.
So the augmentation strategy determines the level of fairness. 
However, it is not easy to control {\it Mixup} by tuning the distribution of the interpolation weight, which is usually a Beta distribution.  
According to \citet{zhang2021does}, 
we consider adversarial training with random perturbation size to augment sensitive features more related to the opposite sensitive attribute.
The perturbation budget of the adversary becomes a key hyperparameter that monitors the generation procedure and controls the degree of fairness.
When the perturbation is significant, the augmented features can be classified into the opposite group with a high probability. 
Then the classifier learnt with large perturbations becomes fairer against the sensitive attribute.

A potential competitive edge of our approach is that we can audit a learnt model at the individual level based on objective criteria: whether the task model is robust to the perturbation of the attribute classifier.
Given an individual, the key problem for testing discrimination is to simulate the corresponding individual with a different protected attribute.
The white-box framework, {\it Counterfactual Fairness} \cite{kusner2017counterfactual} can accurately detect the bias and understand how the model discriminates.
But the reliance on a known causal mechanism limits its application scenarios and may fail to identify instances of legally actionable discrimination.
Some black-box techniques generate mirror individuals via a learnt generative model, e.g. {\it FlipTest} \cite{black2020fliptest}.
Notice that central to assessing fair data generation is a learnt predictor for the sensitive attribute.
However, an intuitive question arises: had an individual been of an opposite sensitive attribute, would the attribute predictor output the opposite attribute?
Without a white-box assumption, the optimal Bayesian classifier may fail to predict the opposite attribute for some individuals, and then the audits run the risk of becoming circular verification. 
The proposed method does not suffer from this problem due to the controllable augmentation.
The perturbation budget during training presents a quantitative standard to perform the model audit.

We summarize the contributions as follows:
\begin{itemize}
    \item We propose Fair-CDA that precisely applies augmentation for sensitive features to achieve fairness while compromising little accuracy. 
    \item The proposed augmentation is  controllable by tuning the perturbation budget of the adversary and provides an audit criterion via adversarial robustness.
    \item Extensive experiments show that Fair-CDA significantly outperforms state-of-the-art methods in various settings, which is effective and  scalable. For instance, Fair-CDA can be applied to different backbones, different tasks, and real application scenarios.
    

\end{itemize}

\section{Preliminaries}\label{sec:definition}

Suppose that data points $\{(x_i,y_i)\}$ are drawn according to some unknown joint distribution over $\cX \times \cY$ with $\cX \subseteq \R^d$. Let $A$ denote the given sensitive attribute that should not be treated differently in decision-making.
Without loss of generality, we consider a binary classification task $\cY = \{0,1\}$ with a binary sensitive attribute $A \in \{0,1\}.$
Let $\hat{Y}$ be the predictor, a random variable that is produced by a model $f:\cX  \rightarrow [0,1]$ as a prediction of $Y$.
In this work, we focus on two widely-used group fairness constraints: Demographic Parity (DP)~\cite{dwork2011fairness} and Equalized Odds (EO)~\cite{hardt2016equality}.

\noindent\textbf{Demographic Parity.} A predictor $\hat{Y}$ satisfies demographic parity (DP) if 
$
P(\hat{Y}\mid A=0)=P(\hat{Y} \mid A=1). 
$
DP requires $\hat{Y}$ to be statistically independent of $A$. In real-world applications, we use DP as the definition to ensure fairness when historically biased decisions may have affected the quality of the collected data and we want to see minority groups receiving positive decisions at the same rate. To evaluate the fairness of a trained model $f$ under this definition, we use the following relaxed metric~\cite{madras2018learning,chuang2021fair}:
\begin{small}
\begin{equation*}
\Delta_{DP}(f) = \big| \mathbb{E}[f(X)|A=0] - \mathbb{E} [f(X)|A=1]  \big|.
\end{equation*}
\end{small}To fulfill the requirement of DP, the evaluation metric $\Delta_{DP}(f)$ shall go to zero.
Since for certain predictions, such as hobbies or expertise, there are indeed differences between groups, meeting the mandatory requirements of DP will greatly decrease the prediction accuracy. Hence, the following alternate criterion is proposed to overcome the limitations of DP.

\noindent\textbf{Equalized Odds.} A predictor $\hat{Y}$ satisfies equalized odds (EO) if 
$
P(\hat{Y} \mid A=0, Y=y)=P(\hat{Y} \mid A=1,Y=y)
$,
for any $y\in \{0,1\}$. EO requires $\hat{Y}$ to be independent of $A$ conditioned on $Y$. In real-world applications, we use EO as a criterion to ensure fairness if there are strict requirements for making correct predictions and we strongly care about the qualifications of candidates when making decisions. Similarly, to evaluate the fairness of a trained model $f$ under EO, we use the following  metric~\cite{madras2018learning,chuang2021fair}:
\begin{small}
\benrr
\Delta_{EO}(f) &=& \sum_{y\in \{0,1\}} \Big| \mathbb{E}[f(X)|A=0, Y=y] \\
&& - \mathbb{E}[f(X)|A=1, Y=y]  \Big|.
\eenrr
\end{small}Different from DP, EO considers the possible correlation between $Y$ and $A$, it does not rule out the perfect predictor even when the base rates differ across groups.

\section{Fair-CDA}

To fulfill the fairness constraint, we shall reduce the dependence of model predictions on sensitive attributes.
Simply removing the sensitive attributes from the inputs does not necessarily lead to a non-discriminatory model prediction, as other attributes in the inputs might encode information for inferring the sensitive attributes~\cite{dwork2011fairness,feldman2015certifying}. Hence, we need to decompose the representations of the inputs into sensitive and non-sensitive features. Sensitive features encode information that can identify whether the inputs belong to a certain group determined by the sensitive attributes, while non-sensitive features retain as much other information as possible~\cite{Zemellearningfair}. Moreover, we shall obfuscate the sensitive features to obtain a fair model. To decompose the high-level representations of the inputs, we train a task model to predict both data labels and sensitive attributes with an orthogonality constraint on gradients for the intermediate features~\cite{bai2020decaug}.

\noindent\textbf{Feature Disentanglement.}
Consider a task with training data $\{(x_i,y_i,a_i)\}_{i=1}^n.$
To decompose the representations,
we denote three feature extractor: $h$, $h_y$ and $h_a$, and write their output features as
\benrr 
z_i = h(x_i), \quad z_i^y = h_y(z_i), \quad  z_i^a = h_a(z_i).
\eenrr 
The mapping $h$ is a pre-extractor that learns the high-level representations of the input.
Then $h_a$ and $h_y$ are two additional extractors after $h$ to obtain sensitive and non-sensitive features.
The principle here is to enforce $h_y$ to extract features that affect the label prediction loss the most will not affect the sensitive attribute prediction loss and vice versa.
Therefore we design a regularization term as follows:
\benr\label{regularization1}
\beta (\cL_i^y + \cL_i^a +  \cL_i^{\perp}),
\eenr
where $\beta$ is a tuning parameter and
\benrr
&& \cL_i^y := \cL_i^y(h, h_y, g_y) = \ell( g_y(z_i^y), y_i), \\
&& \cL_i^a := \cL_i^a(h, h_a, g_a) = \ell( g_a(z_i^a), a_i),
\eenrr
and
\begin{small}
\benn
\cL^{\perp}_i := \cL^{\perp}_i(h_y, h_a, g_y, g_a) =  \frac{\langle \nabla_{z_i} \cL_i^y, \nabla_{z_i} \cL_i^a \rangle^2}{\|\nabla_{z_i} \cL_i^y\|^2\cdot \| \nabla_{z_i} \cL_i^a \|^2}.  
\eenn
\end{small}Here $g_y$ and $g_a$ are two classifier to predict $y$ and $a$ and $\ell$ is the cross-entropy loss. 
The term $\cL^{\perp}_i$ imposes a constraint on gradient orthogonality to disentangle features.
To estimate the feature extractors and the classifiers, Stage 1 of Fair-CDA (Figure.~\ref{fig:model}) minimizes the objective function:
\benrr
 \frac{1}{n} \sum_{i=1}^{n} \cL_i + \beta ( \cL_i^y + \cL_i^a + \cL_i^{\perp}),
\eenrr
where $\cL_i$ is the loss function of the task model $g$ over the disentangled features:
\benr\label{loss1}
\cL_i:=\cL_i(h, h_y,h_a,g)=\ell( g([z^y_i,z^a_i]), y_i). 
\eenr

\noindent{\bf Semantic Augmentation.} 
In Stage 2 of Fair-CDA, 
we do an intervention on the sensitive features to mitigate unfair biases.
Intuitively, a model satisfies the requirement of the fairness constraint if it can make the same prediction for two samples with different sensitive features but the same other features~\cite{kusner2017counterfactual}. 
We augment the sensitive features along the direction of increasing the attribute prediction loss $\cL_i^a$:
\begin{equation}\label{equ:featureaug}
\tilde{z}^a_i = z^a_i + \alpha_i \ \frac{\nabla_{z_i^a} \ell(g_a(z_i^a), a_i)}{\left\| \nabla_{z_i^a} \ell(g_a(z_i^a), a_i) \right\|},
\end{equation}
where $\alpha_i$ is a perturbation size.
Since the direction of the gradient is the direction in which the loss increases most rapidly, augmentation in this way changes $z_i^a$ to the features corresponding to the other sensitive attribute.

\noindent{\bf Transition Path.} \citet{chuang2021fair} interpolates the transition path between groups via Mixup.
\citet{zhang2021does} proves that the adversarial loss can be bounded above by the Mixup loss.
Therefore we generate the transition path over sensitive features by randomizing the perturbation size. 
In this work, we assume $\alpha_i$ is a random variable follows a uniform distribution over $[0, \lambda].$
Here $\lambda$ is the perturbation budget of (\ref{equ:featureaug}) that controls the strength of the augmentation.
After obtaining the generated sensitive features $\tilde{z}_i^a$, we concatenate them with the non-sensitive features $z_i^y$ and train the task model $g$ with $\{ ([z^y_i, \tilde{z}^a_i], y_i) \}.$ The loss function of $g$ over the augmented features is denoted by
\begin{equation}\label{equ:augoriloss}
\tilde{\cL}_i:=\tilde{\cL}_i(h, h_y,h_a,g)=\ell( g([z^y_i, \tilde{z}^a_i]), y_i). 
\end{equation}
together with the aforementioned losses $\cL^1_i$, $\cL^2_i$ and $\cL^{\perp}_i$.

Different from existing works~\cite{lahoti2019ifair,zafar2017fairness,chuang2021fair}, our method strikes a balance between accuracy and fairness via adjusting the perturbation budget $\lambda$. 
When the perturbation is large, the augmented features can be classified into the opposite group with a high probability. 
Then the classifier learnt with large perturbations becomes fairer against the sensitive attribute.

\noindent\textbf{Imputation Model.}
In Stage 2, we use the labels of the original samples to mark the corresponding augmented features.
This is based on the intuition that a model is non-discriminatory if it makes the same prediction for two samples only differing in sensitive features.
However, for certain predictions, there are indeed correlations between labels and sensitive features. Enforcing the model to meet the mandatory requirement of the fairness constraint and ignoring the possible correlations between labels and sensitive features may decrease the prediction accuracy a lot. To further improve the prediction accuracy, we introduce an imputation model to calibrate the labels of the augmented features.
Specifically,  the Stage 1 solution of the task model, denoted by $\check{g}$, is taken to be the imputation model to label the augmented features: $\check{y}_i = \check{g}([z^y_i, \tilde{z}^a_i]).$ 
The loss of predicting $\check{y}_i$ is denoted by
\benr\label{equ:augimploss}
\check{\cL}_i= \check{\cL}_i(h, h_y, h_a, g)=\ell(g([z^y_i, \tilde{z}^a_i]), \check{y}_i).
\eenr
The task model is then trained to predict both the original labels and the labels given by the imputation model. We formulate the final problem of Fair-CDA as minimizing:
\benr\label{equ:final_objective}
\frac{1}{n} \ \sum_{i=1}^{n} \gamma \tilde{\cL}_i + (1-\gamma) \check{\cL}_i + \beta ( \cL_i^1 + \cL_i^2 + \cL_i^{\perp}),
\eenr
where $\gamma$ is a hyper-parameter balancing $\tilde{\cL}_i$ and $\check{\cL}_i.$
For time-saving, our method initializes with the imputation model to solve the optimization problem in (\ref{equ:final_objective}).

\begin{algorithm}[t]\small
\caption{Fair-CDA: Continuous and Directional Augmentation for Group Fairness}
\label{algo1}
\begin{algorithmic}[1]
\REQUIRE Training data $\{(x_i,y_i,a_i)\}_{i=1}^n$, batch sizes $b$, learning rate $\eta_1$, $\eta_2$, perturbation strength $\lambda$, weights $\gamma$, $\beta$, iteration number $T$, $S$
\ENSURE $\theta =(h,h_y,h_a, g, g_y, g_a)$;\\
\noindent{\bf Stage 1:}
\STATE Initialize $\theta^{(0)} =(h^{(0)},h_y^{(0)},h_a^{(0)}, g^{(0)}, g_y^{(0)}, g_a^{(0)})$;
\FOR {$1 \leq t \leq T$} 
\STATE Sample a batch of training data $\{(x_i,y_i,a_i)\}_{i=1}^b$;
\STATE Compute $\cL_i$, $\cL_i^y$, $\cL_i^a$, and $\cL_i^{\perp}$ according to Eq.~(\ref{regularization1}) and Eq.~(\ref{loss1})
\STATE Update $\theta$\\
$\theta^{(t)} = \theta^{(t-1)} -  \frac{\eta_1}{b} \sum\limits_{i=1}^{b} \nabla_\theta \big(\cL_i + \beta( \cL_i^y + \cL_i^a +  \cL_i^{\perp})\big)$;
\ENDFOR\\
\noindent{\bf Stage 2:}
\FOR {$1 \leq s \leq S$} 
		\STATE Sample a batch of training data $\{(x_i,y_i,a_i)\}_{i=1}^b$;
		\FOR {each $(x_i,y_i,a_i)$} 
		\STATE Compute $z_i^y = h_y^{(T+s-1)}\circ h^{(T+s-1)}(x_i)$ and $z_i^a = h_a^{(T+s-1)}\circ h^{(T+s-1)}(x_i)$;
		\STATE Compute $\cL_i^y$, $\cL_i^a$ and $\cL_i^{\perp}$;
		\STATE Randomly draw $\alpha_i$ according to $U(0,\lambda)$;
		\STATE Generate $\tilde{z}^a_i$ according to Eq.~\eqref{equ:featureaug};
		\STATE Compute $\tilde{\cL}_i$ according to Eq.~\eqref{equ:augoriloss};
		\STATE Impute the label $\tilde{y}_i =g^{(T)}([z^y_i, \tilde{z}^a_i])$;
		\STATE Compute $\check{\cL}_i$ according to Eq.~\eqref{equ:augimploss};
		\ENDFOR
		\STATE Update $\theta$: 
        \begin{equation*}
        \begin{split}
            \theta^{(T+s)} =& \theta^{(T+s-1)} - \frac{\eta_2}{b} \sum\limits_{i=1}^{b} \nabla_\theta \big(\gamma \tilde{\cL}_i + (1-\gamma) \check{\cL}_i \\
            & + \beta ( \cL_i^y +  \cL_i^a + \cL_i^{\perp}) \big);
        \end{split}
        \end{equation*}       
		\ENDFOR
	\end{algorithmic}
\end{algorithm}

\noindent{\bf Summary.} As mentioned above, Fair-CDA balances the prediction accuracy and fairness via adjusting the perturbation strength $\lambda$. 
The algorithm is summarized in Algorithm~\ref{algo1}.
Stage~1 disentangles features and learns the task model with the original training samples. In Stage~2, we fine-tune the task model with the augmented features to achieve fairness.

\begin{figure}[t]
\centering
 \includegraphics[width=\linewidth]{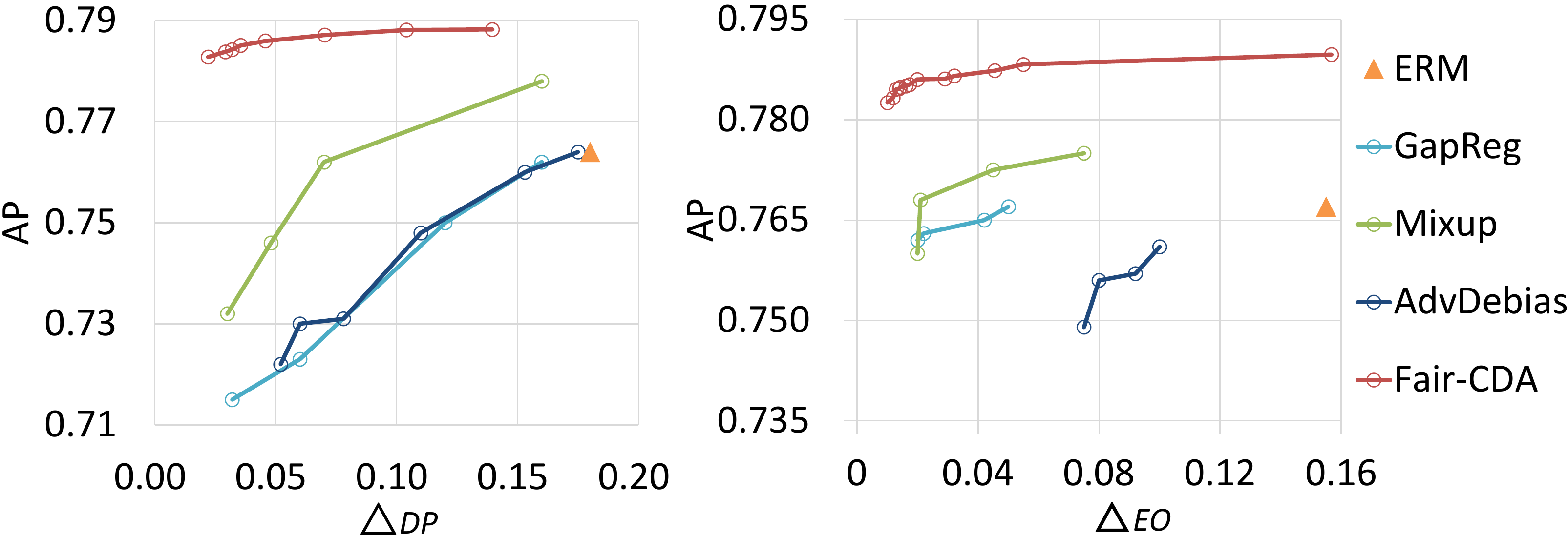}
\caption{
The trade-off between AP and $\Delta_{DP}$ / $\Delta_{EO}$ on Adult dataset.}  
\label{fig:adult_dataset}
\end{figure}

\begin{figure*}[t]
\begin{subfigure}{.32\textwidth}
\centering
\includegraphics[width=.9\linewidth]{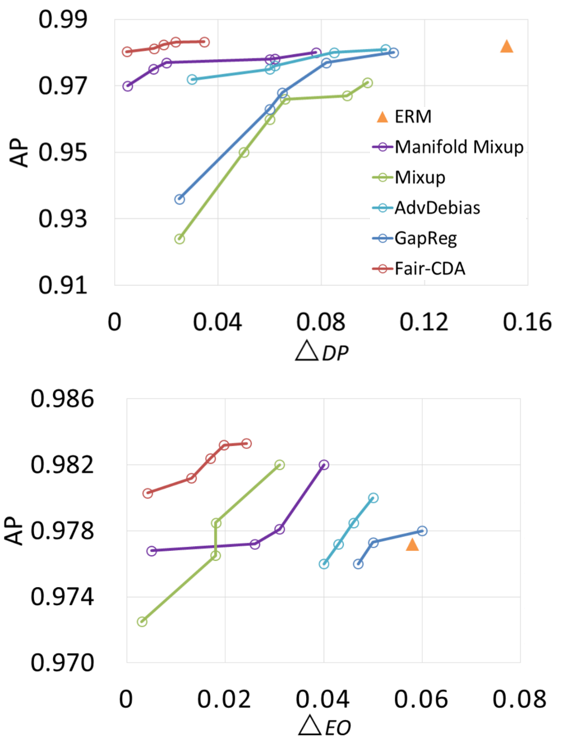}
\caption{Smiling}
\label{fig:celeba_smiling}
\end{subfigure}
\begin{subfigure}{.32\textwidth}
\centering		\includegraphics[width=.9\linewidth]{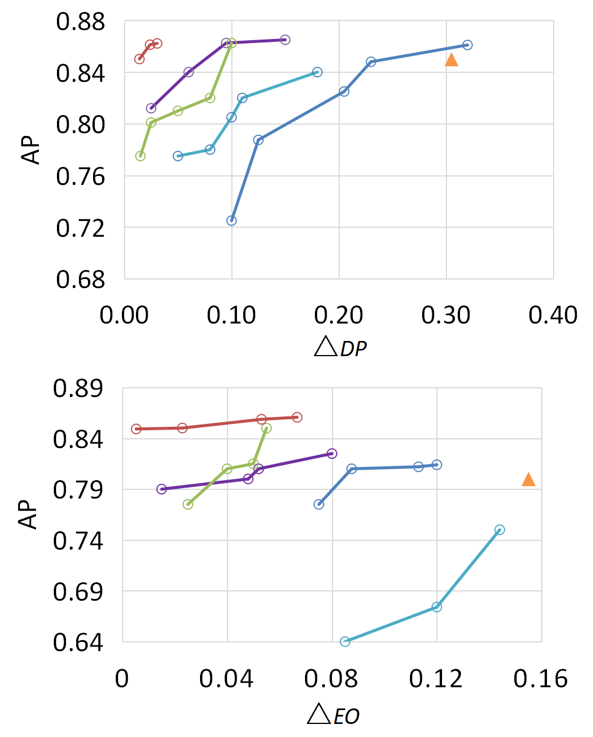}
\caption{Wavy Hair}
\label{fig:celeba_wavy}
\end{subfigure}
\begin{subfigure}{.32\textwidth}
\centering
\includegraphics[width=.9\linewidth]{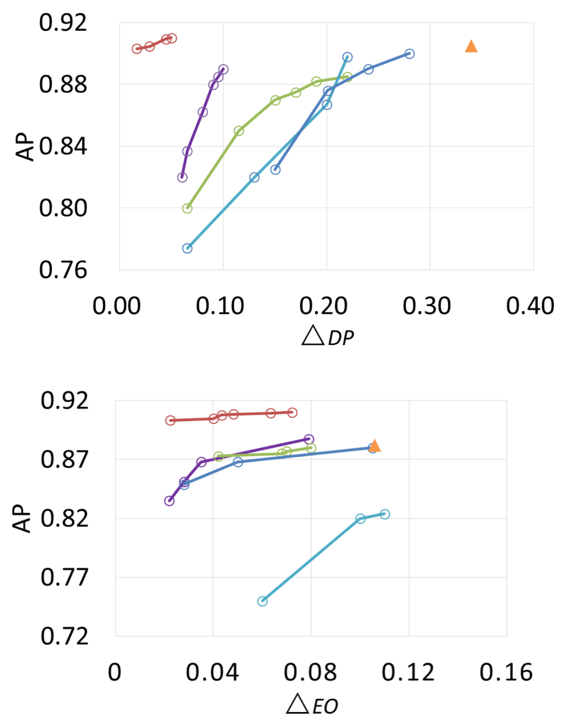}
\caption{Attractive}
\label{fig:celeba_attractive}
\end{subfigure}
\caption{{CelebA.} The trade-off between AP and $\Delta_{DP}$ / $\Delta_{EO}$. Fair-CDA outperforms other methods across tasks.}
\label{fig:celeba_dataset}
\end{figure*}

Our method introduces three additional hyper-parameters: two weights of different losses $\beta$ and $\gamma$, and perturbation budget $\lambda.$ In the experiment, we set $\beta$ according to the initial loss values to make different loss values in the same magnitude range. We adjust $\gamma$ on the Adult dataset~\cite{Dua2017} to get the best accuracy and fairness trade-off on the validation set and then adopt the same value which is 0.9 for all the datasets. Our method balances the prediction accuracy and fairness via adjusting the perturbation strength $\lambda$, while previous works~\cite{chuang2021fair,zhang2018mitigating} balance them via adjusting the weights of the regularization terms. On different datasets, we first conduct experiments with $\lambda=0,1,10,100,1000$ to
narrow down the range of $\lambda$ and then, do grid search between the determined range of $\lambda$ (reported in Appendix) with a budget of 20 points to generate the Pareto Front in Figure~\ref{fig:adult_dataset},\ref{fig:celeba_dataset},\ref{fig:movielens},\ref{fig:ablation_movielens_dataset}. In real-world applications, the number of grid search points can be determined according to the budget.

\section{Experiments on Public Datasets}\label{sec:experiment}

We evaluate Fair-CDA on tabular dataset Adult~\cite{Dua2017}, vision dataset CelebA~\cite{liu2018large}, and recommender dataset MovieLens~\cite{harper2015movielens}. 
We demonstrate the effectiveness of Fair-CDA across diverse tasks and task models. In the ablation studies, we examine the contributions of feature decomposition and
the imputation model.
We compare Fair-CDA with other baseline methods: ERM, GapReg~\cite{chuang2021fair}, AdvDebias~\cite{zhang2018mitigating}, and Mixup / Manifold Mixup~\cite{chuang2021fair} using two metrics: prediction accuracy and fairness. 
To measure the accuracy, we use Average Precision (AP) for tabular (Adult) and vision (CelebA) tasks, and Area Under Curve (AUC) for recommender (MovieLens) task. To measure fairness, we use two widely-used fairness metrics: Demographic Parity (DP) and Equalized Odds (EO) which are defined in Preliminaries. Also, we compare our method with FFAVE and $\beta$-VAE following the setting in~\cite{creager2019flexibly}. Please refer to the Appendix for more details about the datasets.

\noindent\textbf{Unjustified biases from the observed data.} We count the imbalance in the number of training data across sensitive attribute groups and the detailed statistical data are shown in Table~\ref{table:statistics}. In the Adult dataset, the proportion of males with high salaries is significantly higher than that of females. In the CelebA dataset, the proportion of males with a positive label is significantly lower than that of females. In the MovieLens dataset, movies from minority producers also have different positive rates from that of another group. All these imbalances and biases can be inherited and amplified by the models.

\begin{table}[t]
    \centering\small
        \begin{tabular}{llll}
         \hline\hline
        Task & Attribute & Label 
        &Ratio   \\
        \hline
        Adult & Female&	Salary$<=50k$	
        &88.7\%\\
        &Female&	Salary$>50k$ 
        &11.3\%\\
        & Male&	Salary$<=50k$	
        &68.5\%\\        
        & Male&	Salary$>50k$	
        &31.5\%\\         
        \hline
        CelebA & Female& Not Smiling 
        &46.2\%\\
        (Smiling) & Female& Smiling 
        &53.8\%\\
        & Male& Not Smiling 
        &60.1\%\\
        & Male& Smiling 
        &39.9\% \\
        \hline
        CelebA  & Female&	Not Wavy Hair 
        &55.3\%\\
        (Wavy Hair) & Female& Wavy Hair 
        &44.7\%\\
        & Male&	Not Wavy Hair 
        &85.7\%\\
        & Male& Wavy Hair 
        &14.3\%\\ 
        \hline
        CelebA  & Female &	Not Attractive	
        &31.7\%\\
        (Attractive) & Female&Attractive	
        &68.3\%\\
        &Male&Not Attractive	
        &72.1\%\\
        &Male& Attractive	
        &27.9\%\\
        \hline
        MovieLens& Minority & Not Recommend 
        &41.4\%\\
        &Minority &	Recommend 
        &58.6\%\\
        &Majority & Not Recommend 
        &44.1\%\\
        &Majority & Recommend 
        &55.9\% \\
        \hline
        \hline
        \end{tabular}
        \caption{Statistical data of different tasks on three datasets.}
    \label{table:statistics}
    \vspace{-5pt}
\end{table}

\noindent\textbf{Implementations.} Our framework is implemented with PyTorch 1.4 (under BSD license), Python 3.7, and CUDA v9.0. For the baseline methods, we implement with PyTorch 1.3.1 to keep the same setting as their source code. We conducted experiments on NVIDIA Tesla V100. The results of baseline methods on Adult and CelebA are referenced from~\cite{chuang2021fair}, while the results of baseline methods on MovieLens are implemented by ourselves.

\begin{figure}[t]
\centering
\includegraphics[width=1.\linewidth]{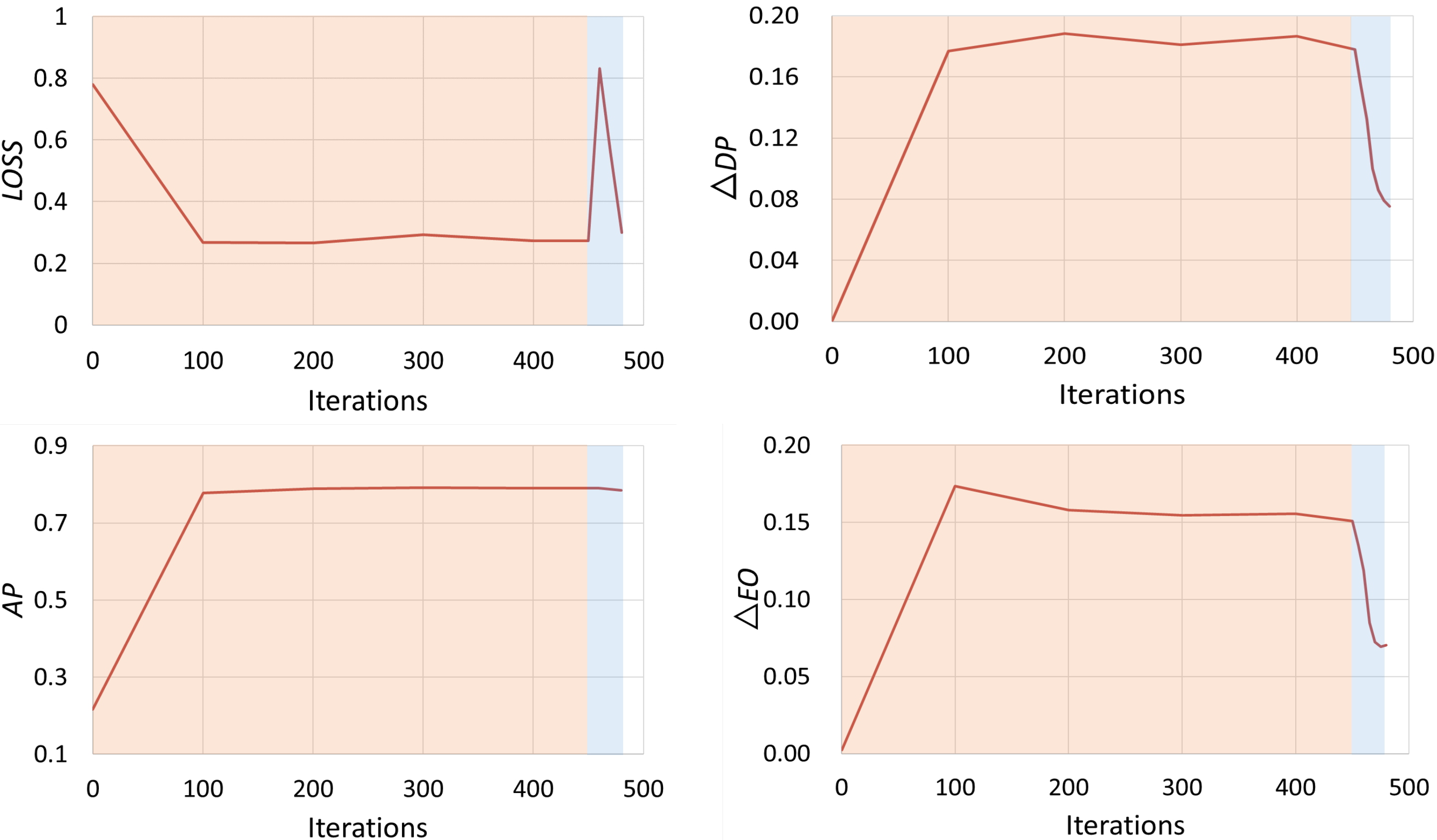}
\caption{The trend of loss and performance during two stages on Adult dataset. The orange color in the figure corresponds to Stage 1, while the blue color represents Stage 2. In Stage 1, Fair-CDA mainly optimizes the prediction accuracy, while in Stage 2, Fair-CDA mainly optimizes the model fairness.}
\label{fig:stage_curve}
\end{figure}

\begin{figure}[t]
\centering
\includegraphics[width=.93\linewidth]{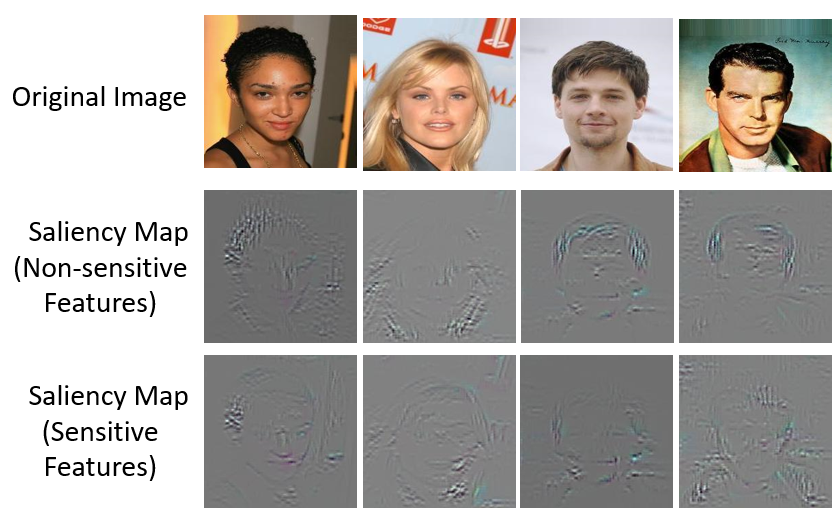}
\caption{Saliency map on the wavy hair recognition task. The saliency maps of sensitive features focus more on the whole face, while those of non-sensitive features focus more on the hair of a man/woman.}
\label{fig:saliency_map}
\end{figure}


\subsection{Results and Discussion}

\noindent\textbf{Results on Adult dataset.} Fair-CDA achieves State-Of-The-Art (SOTA) performance in terms of both fairness and accuracy on the Adult dataset, as shown in Figure~\ref{fig:adult_dataset}. ERM has a moderate AP but poor fairness, while GapReg achieves better fairness but lower AP than ERM. It utilizes the fairness constraint in the training phase, which lacks generalization at evaluation time. 
Fair Mixup achieves a better trade-off compared to the previous three methods but is dominated by Fair-CDA. In particular, Fair-CDA is the only method consistently achieving a higher AP than ERM under two fairness constraints.

To evaluate the feature augmentation, we sample 1,000 training samples from the Adult dataset, extract the sensitive features with the trained model, and generate the augmented features.  
The trained gender classifier, whose prediction accuracy is 86.8\% when using the original sensitive features, predicts opposite labels for all the augmented features.
This means the augmentation policy successfully generates the features corresponding to the opposite sensitive attribute.

\begin{figure}[t]
	\centering
	\includegraphics[width=.98\linewidth]{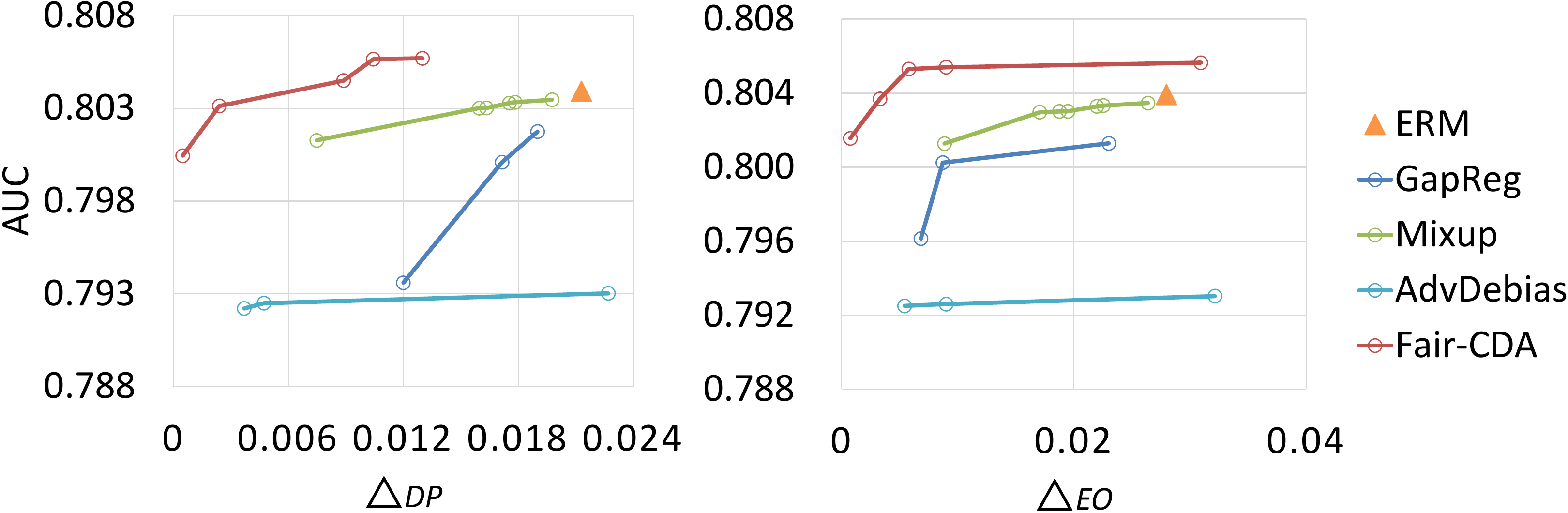}
	\caption{{MovieLens.} The trade-off between AP and $\Delta_{DP}$ / $\Delta_{EO}$. 
	Fair-CDA can reach the smallest $\Delta_{DP}$ / $\Delta_{EO}$ among all the methods without obvious accuracy degradation.}
	\label{fig:movielens}
\end{figure}

To evaluate the prediction accuracy of the imputation model on the augmented samples, we select all the pairs of training samples (208 pairs in total) with the same other attributes but different sensitive attributes and different labels. Intuitively, an accurate imputation model should predict the opposite label for the augmented samples since every augmented sample has a real sample with an opposite label corresponding to it. The imputation model predicts the opposite label on 258 augmented samples (out of 416 samples), which means the prediction accuracy of the task model can be improved with label calibration.

To better understand the training process of Fair-CDA, we plot the trend of loss and performance during two stages on the Adult dataset, as shown in Figure~\ref{fig:stage_curve}. Stage 1 stands for the process of the model trained with original data, while Stage 2 stands for that of the model trained with augmented data. 
At the beginning of Stage 1, the model is initialized with random parameters (without pre-training). The mean value
of the output of both groups tends to be very close, resulting in low fairness disparity. As the training process goes
on, the loss of Fair-CDA converges gradually with the AP rising to a high point. Both DP and EO reach a high value (which means poor fairness). In Stage 2, the AP remains stable, the loss fluctuates at the beginning, and finally drops to a low point. Both DP and EO reach a low value. 

\begin{figure}[t]
	\centering
\includegraphics[width=.99\linewidth]{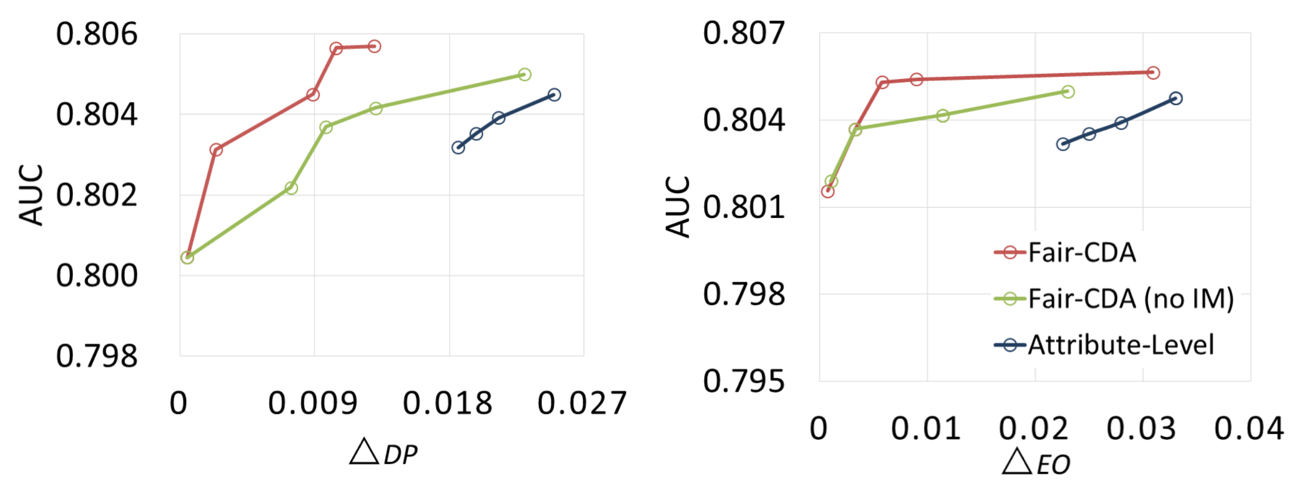}
	\caption{Ablation studies on MovieLens dataset. 
	Fair-CDA without imputation model (Fair-CDA (no IM)) can still satisfy the fairness requirement but suffer an accuracy loss. By generating the samples with opposite sensitive attributes (Attribute-Level), the unfairness can hardly be decreased.
	}
\label{fig:ablation_movielens_dataset}
\end{figure}

\noindent\textbf{Results on CelebA.} To illustrate each model’s ability for vision tasks, we choose smiling, wavy hair, and attractive to form three binary classification tasks. As shown in Figure~\ref{fig:celeba_dataset}, Fair-CDA achieves SOTA performance followed by two mixup methods. It is worth mentioning that the DP and EO gap of these methods on the smiling recognition task is smaller compared with other tasks, which is a relatively fair scenario, but Fair-CDA can still improve the fairness. 
Also, Fair-CDA is the only method that achieves considerable accuracy given high fairness requirements on both tasks. 

To visualize the effect of feature decomposition, we adopt deep neural network
interpretability methods in~\cite{adebayo2018sanity}.
We draw the saliency map on the wavy hair recognition task, as shown in Figure~\ref{fig:saliency_map}. Sensitive features are those strongly related to gender, while non-sensitive features are those strongly related to wavy hair. It can be seen that the saliency maps of sensitive features focus more on the whole face, while those of non-sensitive features focus more on the hair of a man/woman. 

Additionally, we evaluate Fair-CDA on more sensitive features on CelebA dataset. We implement Fair-CDA on the same task as that in~\cite{creager2019flexibly} (CelebA Heavy-Makeup recognition task) to compare our method with two VAE methods. Noted that the fairness metric is demographic parity and the performance metric is accuracy in this setting. As shown in Table~\ref{table:more_attribute}, Fair-CDA outperforms FFVAE and $\beta$-VAE on CelebA Heavy-Makeup recognition task considering three different sensitive attributes.

\begin{table}[!t]
    \centering\small
    \begin{tabular}{llll}
    \hline
    \hline
         & Male  & Chubby &  Eyeglasses \\ 
        ~ & $\Delta_{DP}$/Acc  & $\Delta_{DP}$/Acc  & $\Delta_{DP}$/Acc  \\ \hline
         $\beta$-VAE & 0.330/0.712 & 0.202/0.732 & 0.250/0.715  \\ 
        ~ & 0.400/0.725 & 0.220/0.740 & 0.280/0.735  \\ \hline
        FFVAE & 0.330/0.730 & 0.202/0.748 & 0.250/0.725  \\ 
        ~ & 0.400/0.752 & 0.400/0.825 & 0.400/0.824  \\ \hline
        Fair-CDA & 0.234/0.733 & 0.184/0.816 & 0.217/0.814  \\
        ~ & 0.369/0.836 & 0.197/0.825 & 0.245/0.824  \\ \hline\hline
    \end{tabular}
    \caption{Results on CelebA dataset. Compared with two VAE methods, Fair-CDA improves the fairness measurement $\Delta_{DP}$ and accuracy significantly.}
    \label{table:more_attribute}
\end{table}

\noindent\textbf{Results on MovieLens dataset.}
Recommendation, a common scenario of machine learning, poses unique challenges for applying fairness and non-discrimination concepts. We choose the rating recognition task on MovieLens to evaluate different fair methods. Similar to the trends on the Adult dataset, Fair-CDA achieves SOTA performance
followed by Fair Mixup and GapReg, as shown in Figure~\ref{fig:movielens}. AdvDebias achieves better fairness than ERM accompanied by severe accuracy degradation. In addition, Fair-CDA can reach the smallest $\Delta_{DP}$ and $\Delta_{EO}$ among all the methods, which shows its superior ability to obtain the group fairness.

\subsection{Ablation Studies}

\noindent\textbf{Without imputation model.} To examine whether the imputation model contributes to performance, we train Fair-CDA without imputation model (Fair-CDA (no IM)) on MovieLens dataset and plot the Pareto Front in Figure~\ref{fig:ablation_movielens_dataset}. We can see the Pareto Front of Fair-CDA dominates that of Fair-CDA (no IM) for both DP and EO. Without an imputation model, Fair-CDA can still achieve good fairness but suffer a little accuracy loss. 

\noindent\textbf{Sample generating at the attribute level.} To illustrate the effectiveness of feature decomposition, we use a naive way to generate the flip sample. Simply flipping the value of the sensitive attribute with a specific probability during the training phase, we can get the results of model training with different data distributions. We name the method as Attribute-Level. By setting different probabilities, we can get the Pareto Front, as shown in Figure~\ref{fig:ablation_movielens_dataset}. Compared with ERM, Attribute-Level can mitigate the unfairness to some extent, while it can not solve the problem mentioned earlier: other variables correlated with sensitive attributes can serve as a source for unfairness.

\section{Experiments on Product.}

To further validate our algorithm in realistic scenarios, we deploy Fair-CDA in an online course recommender system. There are nearly 100,000 users and more than 12,000 courses developed by more than 100 different suppliers, nearly 50\% of the courses coming from top 5\% suppliers. Thus we consider supplier as the sensitive attribute to evaluate fairness. Similar to the previous setting on MovieLens, we divide the suppliers into the majority and minority groups according to the number of courses developed by the suppliers. The top 5\% suppliers who provide nearly 50\% of the courses are regarded as the majority supplier and the remaining suppliers are regarded as the minority supplier. In this scenario, we choose Equalized Odds as the fairness measurement since it has been shown that demographic parity causes a loss in the utility and infringes individual fairness~\cite{singh2018fairness}, and we adopt AUC and Top-10 Recall as the offline accuracy 
evaluation. We use LightGCN~\cite{LightGCN2020} as the backbone network and compare Fair-CDA with the original LightGCN method. The results are shown in Table~\ref{table:offline}. Fair-CDA achieves better performance on both accuracy and fairness measurements than the baseline method.

\begin{table}[t]
    \centering\small
        \resizebox{0.47\textwidth}{!}{
        \begin{tabular}{lccc}
        \hline\hline
        Method & AUC&  $\Delta_{EO}$ &  Top-10 Recall  \\
        \hline
        LightGCN (Baseline)  & 0.9503 &  0.0448  & 0.1116   \\
        \hline
        Fair-CDA & 0.9679    & 0.0227 & 0.1328 \\
        \hline\hline
        \end{tabular}
        }
        \caption{Offline results on a product dataset from an online course recommender system.} 
    \label{table:offline}
    \vspace{-5pt}
\end{table}

Inspired by the performance of offline evaluation, we implement and deploy Fair-CDA in the production environment and verify its effectiveness through a consecutive online A/B test. We split the users into two groups uniformly, each of which has an average of 3000 users every week. The first group gets courses recommended by the baseline model, and the Fair-CDA generates recommendations for the other group. The two models are updated daily. After a 5-week online A/B test, the Fair-CDA is consistently superior than the baseline model, with an average Click Through Rate (CTR) improvement of 6.5\%. During the online A/B test, our method increases the diversity of recommended courses and enhances group fairness, resulting in a higher CTR.

\section{Conclusions}

We propose {\it Fair-CDA} to counter the unfairness problem via feature decomposition and data augmentation. Fair-CDA improves fairness and minimizes the impact on accuracy. We experimentally compare our method with other state-of-the-art fairness methods on various benchmarks and show that Fair-CDA significantly outperforms the other methods in all the experimental settings.


\section*{Acknowledgements}

This work was partially supported by JCYJ20220530143600001,
by the Basic Research Project No. HZQB-KCZYZ-2021067 of Hetao Shenzhen HK S\&T Cooperation Zone, by the National Key R\&D Program of China with grant No.2018YFB1800800, by SGDX20211123112401002, by Shenzhen Outstanding Talents Training Fund, by Guangdong Research Project No. 2017ZT07X152 and No. 2019CX01X104, by the Guangdong Provincial Key Laboratory of Future Networks of Intelligence (Grant No. 2022B1212010001), by the NSFC 61931024\&8192 2046, by NSFC-Youth 62106154, by zelixir biotechnology company Fund, by Tencent Open Fund, and by ITSO at CUHKSZ. Chuanlong Xie was partially supported by NSFC No.12201048 and the Interdisciplinary Intelligence SuperComputer Center of Beijing Normal University at Zhuhai.

\nocite{*}
\bibliography{aaai23}

\begin{thebibliography}{55}
\providecommand{\natexlab}[1]{#1}

\bibitem[{Adebayo et~al.(2018)Adebayo, Gilmer, Muelly, Goodfellow, Hardt, and
  Kim}]{adebayo2018sanity}
Adebayo, J.; Gilmer, J.; Muelly, M.; Goodfellow, I.; Hardt, M.; and Kim, B.
  2018.
\newblock Sanity checks for saliency maps.
\newblock \emph{arXiv preprint arXiv:1810.03292}.

\bibitem[{Agarwal et~al.(2018)Agarwal, Beygelzimer, Dud{\'\i}k, Langford, and
  Wallach}]{agarwal2018reductions}
Agarwal, A.; Beygelzimer, A.; Dud{\'\i}k, M.; Langford, J.; and Wallach, H.
  2018.
\newblock A reductions approach to fair classification.
\newblock In \emph{International Conference on Machine Learning}, 60--69. PMLR.

\bibitem[{Arjovsky et~al.(2019)Arjovsky, Bottou, Gulrajani, and
  Lopez-Paz}]{arjovsky2019invariant}
Arjovsky, M.; Bottou, L.; Gulrajani, I.; and Lopez-Paz, D. 2019.
\newblock Invariant Risk Minimization.
\newblock \emph{arXiv:1907.02893}.

\bibitem[{Bai et~al.(2020)Bai, Sun, Hong, Zhou, Ye, Ye, Chan, and
  Li}]{bai2020decaug}
Bai, H.; Sun, R.; Hong, L.; Zhou, F.; Ye, N.; Ye, H.-J.; Chan, S.-H.~G.; and
  Li, Z. 2020.
\newblock DecAug: Out-of-Distribution Generalization via Decomposed Feature
  Representation and Semantic Augmentation.
\newblock \emph{arXiv preprint arXiv:2012.09382}.

\bibitem[{Barocas, Hardt, and Narayanan(2019)}]{barocas}
Barocas, S.; Hardt, M.; and Narayanan, A. 2019.
\newblock \emph{Fairness and Machine Learning: Limitations and Opportunities}.
\newblock fairmlbook.org.
\newblock Accessed: 2019.

\bibitem[{Beutel et~al.(2019)Beutel, Chen, Doshi, Qian, Wei, Wu, Heldt, Zhao,
  Hong, Chi et~al.}]{beutel2019fairness}
Beutel, A.; Chen, J.; Doshi, T.; Qian, H.; Wei, L.; Wu, Y.; Heldt, L.; Zhao,
  Z.; Hong, L.; Chi, E.~H.; et~al. 2019.
\newblock Fairness in recommendation ranking through pairwise comparisons.
\newblock In \emph{Proceedings of the 25th ACM SIGKDD International Conference
  on Knowledge Discovery \& Data Mining}, 2212--2220.

\bibitem[{Black, Yeom, and Fredrikson(2020)}]{black2020fliptest}
Black, E.; Yeom, S.; and Fredrikson, M. 2020.
\newblock Fliptest: fairness testing via optimal transport.
\newblock In \emph{Proceedings of the 2020 Conference on Fairness,
  Accountability, and Transparency}, 111--121.

\bibitem[{Bose and Hamilton(2019)}]{bose2019compositional}
Bose, A.; and Hamilton, W. 2019.
\newblock Compositional fairness constraints for graph embeddings.
\newblock In \emph{International Conference on Machine Learning}, 715--724.
  PMLR.

\bibitem[{Calmon et~al.(2017)Calmon, Wei, Vinzamuri, Ramamurthy, and
  Varshney}]{calmon2017optimized}
Calmon, F.~P.; Wei, D.; Vinzamuri, B.; Ramamurthy, K.~N.; and Varshney, K.~R.
  2017.
\newblock Optimized pre-processing for discrimination prevention.
\newblock In \emph{Proceedings of the 31st International Conference on Neural
  Information Processing Systems}, 3995--4004.

\bibitem[{Chuang and Mroueh(2021)}]{chuang2021fair}
Chuang, C.-Y.; and Mroueh, Y. 2021.
\newblock Fair Mixup: Fairness via Interpolation.
\newblock In \emph{International Conference on Learning Representations}.

\bibitem[{Cotter et~al.(2019)Cotter, Gupta, Jiang, Srebro, Sridharan, Wang,
  Woodworth, and You}]{cotter2019training}
Cotter, A.; Gupta, M.; Jiang, H.; Srebro, N.; Sridharan, K.; Wang, S.;
  Woodworth, B.; and You, S. 2019.
\newblock Training well-generalizing classifiers for fairness metrics and other
  data-dependent constraints.
\newblock In \emph{International Conference on Machine Learning}, 1397--1405.
  PMLR.

\bibitem[{Creager et~al.(2019)Creager, Madras, Jacobsen, Weis, Swersky,
  Pitassi, and Zemel}]{creager2019flexibly}
Creager, E.; Madras, D.; Jacobsen, J.-H.; Weis, M.; Swersky, K.; Pitassi, T.;
  and Zemel, R. 2019.
\newblock Flexibly Fair Representation Learning by Disentanglement.
\newblock In \emph{International conference on machine learning}, 1436--1445.
  PMLR.

\bibitem[{Dou et~al.(2019)Dou, Castro, Kamnitsas, and Glocker}]{dou2019domain}
Dou, Q.; Castro, D.~C.; Kamnitsas, K.; and Glocker, B. 2019.
\newblock Domain Generalization via Model-Agnostic Learning of Semantic
  Features.
\newblock In \emph{Advances in Neural Information Processing Systems}.

\bibitem[{Dua and Graff(2017)}]{Dua2017}
Dua, D.; and Graff, C. 2017.
\newblock {UCI} Machine Learning Repository.
\newblock \url{http://archive.ics.uci.edu/ml}.
\newblock Accessed: 2017.

\bibitem[{Dwork et~al.(2012)Dwork, Hardt, Pitassi, Reingold, and
  Zemel}]{dwork2011fairness}
Dwork, C.; Hardt, M.; Pitassi, T.; Reingold, O.; and Zemel, R. 2012.
\newblock Fairness through awareness.
\newblock In \emph{Proceedings of the 3rd innovations in theoretical computer
  science conference}, 214--226.

\bibitem[{Edwards and Storkey(2015)}]{edwards2015censoring}
Edwards, H.; and Storkey, A. 2015.
\newblock Censoring representations with an adversary.
\newblock \emph{arXiv preprint arXiv:1511.05897}.

\bibitem[{Ekstrand et~al.(2021)Ekstrand, Das, Burke, and
  Diaz}]{ekstrand2021fairness}
Ekstrand, M.~D.; Das, A.; Burke, R.; and Diaz, F. 2021.
\newblock Fairness and Discrimination in Information Access Systems.
\newblock \emph{arXiv preprint arXiv:2105.05779}.

\bibitem[{Feldman et~al.(2015)Feldman, Friedler, Moeller, Scheidegger, and
  Venkatasubramanian}]{feldman2015certifying}
Feldman, M.; Friedler, S.~A.; Moeller, J.; Scheidegger, C.; and
  Venkatasubramanian, S. 2015.
\newblock Certifying and removing disparate impact.
\newblock In \emph{proceedings of the 21th ACM SIGKDD international conference
  on knowledge discovery and data mining}, 259--268.

\bibitem[{Feng et~al.(2019)Feng, Yang, Lyu, Tan, Sun, and
  Wang}]{feng2019learning}
Feng, R.; Yang, Y.; Lyu, Y.; Tan, C.; Sun, Y.; and Wang, C. 2019.
\newblock Learning fair representations via an adversarial framework.
\newblock \emph{arXiv preprint arXiv:1904.13341}.

\bibitem[{Goodfellow et~al.(2014)Goodfellow, Pouget-Abadie, Mirza, Xu,
  Warde-Farley, Ozair, Courville, and Bengio}]{goodfellow2014generative}
Goodfellow, I.; Pouget-Abadie, J.; Mirza, M.; Xu, B.; Warde-Farley, D.; Ozair,
  S.; Courville, A.; and Bengio, Y. 2014.
\newblock Generative adversarial nets.
\newblock \emph{Advances in neural information processing systems}, 27.

\bibitem[{Hardt, Price, and Srebro(2016)}]{hardt2016equality}
Hardt, M.; Price, E.; and Srebro, N. 2016.
\newblock Equality of opportunity in supervised learning.
\newblock \emph{Advances in neural information processing systems}, 29:
  3315--3323.

\bibitem[{Harper and Konstan(2015)}]{harper2015movielens}
Harper, F.~M.; and Konstan, J.~A. 2015.
\newblock The movielens datasets: History and context.
\newblock \emph{Acm transactions on interactive intelligent systems (tiis)},
  5(4): 1--19.

\bibitem[{He et~al.(2016)He, Zhang, Ren, and Sun}]{he2016deep}
He, K.; Zhang, X.; Ren, S.; and Sun, J. 2016.
\newblock Deep Resibarol Learning for Image Recognition.
\newblock In \emph{Proceedings of the IEEE Conference on Computer Vision and
  Pattern Recognition}, 770--778.

\bibitem[{He et~al.(2020)He, Deng, Wang, Li, Zhang, and Wang}]{LightGCN2020}
He, X.; Deng, K.; Wang, X.; Li, Y.; Zhang, Y.; and Wang, M. 2020.
\newblock LightGCN: Simplifying and Powering Graph Convolution Network for
  Recommendation.
\newblock In \emph{Proceedings of the 43rd International ACM SIGIR Conference
  on Research and Development in Information Retrieval}, SIGIR '20, 639–648.

\bibitem[{Jha, Vinzamuri, and Reddy(2021)}]{jha2021fair}
Jha, A.; Vinzamuri, B.; and Reddy, C.~K. 2021.
\newblock Fair Representation Learning using Interpolation Enabled
  Disentanglement.
\newblock \emph{arXiv preprint arXiv:2108.00295}.

\bibitem[{{Joachims} and {Swaminathan}(2016)}]{joachims2016counterfactual}
{Joachims}, T.; and {Swaminathan}, A. 2016.
\newblock Counterfactual Evaluation and Learning for Search, Recommendation and
  Ad Placement.
\newblock In \emph{Proceedings of the 39th International ACM SIGIR conference
  on Research and Development in Information Retrieval}, 1199--1201.

\bibitem[{Kamiran and Calders(2009)}]{2009Classifying}
Kamiran, F.; and Calders, T. 2009.
\newblock Classifying without discriminating.
\newblock In \emph{International Conference on Computer}.

\bibitem[{Kamiran and Calders(2012)}]{Kamiran2012}
Kamiran, F.; and Calders, T. 2012.
\newblock Data preprocessing techniques for classification without
  discrimination.
\newblock \emph{Knowledge and Information Systems}, 33(1): 1--33.

\bibitem[{Khademi et~al.(2019)Khademi, Lee, Foley, and
  Honavar}]{khademi2019fairness}
Khademi, A.; Lee, S.; Foley, D.; and Honavar, V. 2019.
\newblock Fairness in algorithmic decision making: An excursion through the
  lens of causality.
\newblock In \emph{The World Wide Web Conference}, 2907--2914.

\bibitem[{Kuang et~al.(2018)Kuang, Cui, Athey, Xiong, and Li}]{Kuang2018}
Kuang, K.; Cui, P.; Athey, S.; Xiong, R.; and Li, B. 2018.
\newblock Stable Prediction across Unknown Environments.
\newblock In \emph{Proceedings of the 24th ACM SIGKDD International Conference
  on Knowledge Discovery \& Data Mining}, 1617--1626.

\bibitem[{Kusner et~al.(2017)Kusner, Loftus, Russell, and
  Silva}]{kusner2017counterfactual}
Kusner, M.~J.; Loftus, J.; Russell, C.; and Silva, R. 2017.
\newblock Counterfactual fairness.
\newblock \emph{Advances in neural information processing systems}, 30.

\bibitem[{Lahoti, Gummadi, and Weikum(2019)}]{lahoti2019ifair}
Lahoti, P.; Gummadi, K.~P.; and Weikum, G. 2019.
\newblock ifair: Learning individu ally fair data representations for
  algorithmic decision making.
\newblock In \emph{2019 IEEE 35th International Conference on Data Engineering
  (ICDE)}, 1334--1345. IEEE.

\bibitem[{Li et~al.(2022)Li, Xu, Lai, and Gu}]{li2022towards}
Li, Y.; Xu, K.; Lai, R.; and Gu, L. 2022.
\newblock Towards an Effective Orthogonal Dictionary Convolution Strategy.
\newblock In \emph{Proceedings of the 36th AAAI Conference on Artificial
  Intelligence (AAAI)}, volume~36, 1473--1481.

\bibitem[{Liu et~al.(2018)Liu, Luo, Wang, and Tang}]{liu2018large}
Liu, Z.; Luo, P.; Wang, X.; and Tang, X. 2018.
\newblock Large-scale celebfaces attributes (celeba) dataset.
\newblock \emph{Retrieved August}, 15(2018): 11.

\bibitem[{Locatello et~al.(2019)Locatello, Abbati, Rainforth, Bauer,
  Sch{\"o}lkopf, and Bachem}]{locatello2019fairness}
Locatello, F.; Abbati, G.; Rainforth, T.; Bauer, S.; Sch{\"o}lkopf, B.; and
  Bachem, O. 2019.
\newblock On the Fairness of Disentangled Representations.
\newblock \emph{Advances in Neural Information Processing Systems}, 32:
  14611--14624.

\bibitem[{Louizos et~al.(2015)Louizos, Swersky, Li, Welling, and
  Zemel}]{louizos2015variational}
Louizos, C.; Swersky, K.; Li, Y.; Welling, M.; and Zemel, R. 2015.
\newblock The variational fair autoencoder.
\newblock \emph{arXiv preprint arXiv:1511.00830}.

\bibitem[{Madras et~al.(2018)Madras, Creager, Pitassi, and
  Zemel}]{madras2018learning}
Madras, D.; Creager, E.; Pitassi, T.; and Zemel, R. 2018.
\newblock Learning adversarially fair and transferable representations.
\newblock In \emph{International Conference on Machine Learning}, 3384--3393.
  PMLR.

\bibitem[{McNamara, Ong, and Williamson(2017)}]{mcnamara2017provably}
McNamara, D.; Ong, C.~S.; and Williamson, R.~C. 2017.
\newblock Provably fair representations.
\newblock \emph{arXiv preprint arXiv:1710.04394}.

\bibitem[{McNamara, Ong, and Williamson(2019)}]{mcnamara2019costs}
McNamara, D.; Ong, C.~S.; and Williamson, R.~C. 2019.
\newblock Costs and benefits of fair representation learning.
\newblock In \emph{Proceedings of the 2019 AAAI/ACM Conference on AI, Ethics,
  and Society}, 263--270.

\bibitem[{Morik et~al.(2020)Morik, Singh, Hong, and Joachims}]{Morik_2020}
Morik, M.; Singh, A.; Hong, J.; and Joachims, T. 2020.
\newblock Controlling Fairness and Bias in Dynamic Learning-to-Rank.
\newblock \emph{Proceedings of the 43rd International ACM SIGIR Conference on
  Research and Development in Information Retrieval}.

\bibitem[{Park et~al.(2021)Park, Hwang, Kim, and Byun}]{park2021learning}
Park, S.; Hwang, S.; Kim, D.; and Byun, H. 2021.
\newblock Learning Disentangled Representation for Fair Facial Attribute
  Classification via Fairness-aware Information Alignment.
\newblock In \emph{Proceedings of the AAAI Conference on Artificial
  Intelligence}, volume~35, 2403--2411.

\bibitem[{Quadrianto, Sharmanska, and Thomas(2019)}]{quadrianto2019discovering}
Quadrianto, N.; Sharmanska, V.; and Thomas, O. 2019.
\newblock Discovering fair representations in the data domain.
\newblock In \emph{Proceedings of the IEEE/CVF Conference on Computer Vision
  and Pattern Recognition}, 8227--8236.

\bibitem[{Sarhan et~al.(2020)Sarhan, Navab, Eslami, and
  Albarqouni}]{sarhan2020fairness}
Sarhan, M.~H.; Navab, N.; Eslami, A.; and Albarqouni, S. 2020.
\newblock Fairness by learning orthogonal disentangled representations.
\newblock In \emph{Computer Vision--ECCV 2020: 16th European Conference,
  Glasgow, UK, August 23--28, 2020, Proceedings, Part XXIX 16}, 746--761.
  Springer.

\bibitem[{Singh and Joachims(2018)}]{singh2018fairness}
Singh, A.; and Joachims, T. 2018.
\newblock Fairness of exposure in rankings.
\newblock In \emph{Proceedings of the 24th ACM SIGKDD International Conference
  on Knowledge Discovery \& Data Mining}, 2219--2228.

\bibitem[{Song et~al.(2019)Song, Kalluri, Grover, Zhao, and
  Ermon}]{song2019learning}
Song, J.; Kalluri, P.; Grover, A.; Zhao, S.; and Ermon, S. 2019.
\newblock Learning controllable fair representations.
\newblock In \emph{The 22nd International Conference on Artificial Intelligence
  and Statistics}, 2164--2173. PMLR.

\bibitem[{Verma et~al.(2019)Verma, Lamb, Beckham, Najafi, Mitliagkas,
  Lopez-Paz, and Bengio}]{verma2019manifold}
Verma, V.; Lamb, A.; Beckham, C.; Najafi, A.; Mitliagkas, I.; Lopez-Paz, D.;
  and Bengio, Y. 2019.
\newblock Manifold mixup: Better representations by interpolating hidden
  states.
\newblock In \emph{International Conference on Machine Learning}, 6438--6447.
  PMLR.

\bibitem[{Wang et~al.(2017)Wang, Fu, Fu, and Wang}]{wang2017deep}
Wang, R.; Fu, B.; Fu, G.; and Wang, M. 2017.
\newblock Deep \& cross network for ad click predictions.
\newblock In \emph{Proceedings of the ADKDD'17}, 1--7.

\bibitem[{Xu et~al.(2018)Xu, Yuan, Zhang, and Wu}]{xu2018fairgan}
Xu, D.; Yuan, S.; Zhang, L.; and Wu, X. 2018.
\newblock Fairgan: Fairness-aware generative adversarial networks.
\newblock In \emph{2018 IEEE International Conference on Big Data (Big Data)},
  570--575. IEEE.

\bibitem[{Zafar et~al.(2017)Zafar, Valera, Rogriguez, and
  Gummadi}]{zafar2017fairness}
Zafar, M.~B.; Valera, I.; Rogriguez, M.~G.; and Gummadi, K.~P. 2017.
\newblock Fairness constraints: Mechanisms for fair classification.
\newblock In \emph{Artificial Intelligence and Statistics}, 962--970. PMLR.

\bibitem[{Zeiler and Fergus(2014)}]{zeiler2014visualizing}
Zeiler, M.~D.; and Fergus, R. 2014.
\newblock Visualizing and understanding convolutional networks.
\newblock In \emph{European conference on computer vision}, 818--833. Springer.

\bibitem[{Zemel et~al.(2013)Zemel, Wu, Swersky, Pitassi, and
  Dwork}]{Zemellearningfair}
Zemel, R.; Wu, Y.; Swersky, K.; Pitassi, T.; and Dwork, C. 2013.
\newblock Learning Fair Representations.
\newblock In \emph{International Conference on Machine Learning}, 325–333.

\bibitem[{Zhang, Lemoine, and Mitchell(2018)}]{zhang2018mitigating}
Zhang, B.~H.; Lemoine, B.; and Mitchell, M. 2018.
\newblock Mitigating unwanted biases with adversarial learning.
\newblock In \emph{Proceedings of the 2018 AAAI/ACM Conference on AI, Ethics,
  and Society}, 335--340.

\bibitem[{Zhang et~al.(2018)Zhang, Cisse, Dauphin, and
  Lopez-Paz}]{zhang2018mixup}
Zhang, H.; Cisse, M.; Dauphin, Y.~N.; and Lopez-Paz, D. 2018.
\newblock mixup: Beyond Empirical Risk Minimization.
\newblock In \emph{International Conference on Learning Representations}.

\bibitem[{Zhang et~al.(2021)Zhang, Deng, Kawaguchi, Ghorbani, and
  Zou}]{zhang2021does}
Zhang, L.; Deng, Z.; Kawaguchi, K.; Ghorbani, A.; and Zou, J. 2021.
\newblock How Does Mixup Help With Robustness and Generalization?
\newblock In \emph{International Conference on Learning Representations}.

\bibitem[{Zhou et~al.(2021)Zhou, Zhang, Nair, Singhal, Chen, and
  Sudjianto}]{zhou2021bias}
Zhou, N.; Zhang, Z.; Nair, V.~N.; Singhal, H.; Chen, J.; and Sudjianto, A.
  2021.
\newblock Bias, Fairness, and Accountability with AI and ML Algorithms.
\newblock \emph{arXiv preprint arXiv:2105.06558}.

\end{thebibliography}

\end{document}